\documentclass{article}

\usepackage{iclr2021_conference,times}

\usepackage[utf8]{inputenc} 
\usepackage[T1]{fontenc}    
\usepackage{hyperref}       
\usepackage{url}            
\usepackage{booktabs}       
\usepackage{amsfonts}       
\usepackage{nicefrac}       
\usepackage{microtype}      

\usepackage{times}
\usepackage{epsfig}
\usepackage{caption}
\usepackage{graphicx}
\usepackage{amssymb}
\usepackage{bm}
\usepackage{enumerate}
\usepackage{color}
\usepackage{multirow}
\usepackage{array}
\usepackage{bigstrut}
\usepackage{algorithmic}
\usepackage{algorithm}
\usepackage{arydshln}
\usepackage{subcaption}
\usepackage{wrapfig}
\usepackage{comment}
\usepackage{amsmath}
\usepackage{enumitem}

\newcommand{\T}[1]{{\mathcal{#1}}} 
\newcommand{\V}[1]{{\mathbf{#1}}} 

\newcommand{\ours}{\textsc{dyari}}
\newcommand{\nri}{\textsc{nri}}
\newcommand{\inter}{\textsc{in}}

\usepackage{color}

\title{Dynamic   Relational Inference in  \\ Multi-Agent Trajectories}

\author{%
  Ruichao Xiao \thanks{Authors contributed equally}\\
  \texttt{xiao.ruic@northeastern.edu} \\
    {Northeastern University}
  \And
  Manish Kumar Singh \footnotemark[\value{footnote}]\\
  \texttt{mksingh@eng.ucsd.edu} \\
    {UC San Diego}
  \And
  Rose Yu \\
  \texttt{roseyu@ucsd.edu} \\
    {UC San Diego}

}
\iclrfinalcopy
\begin{document}

\maketitle
\begin{abstract}
 Unsupervised learning of  interactions from multi-agent trajectories has broad applications in physics, vision and robotics.  However, existing  neural relational inference works are limited to \textit{static} relations. In this paper, we consider a more general setting  of  dynamic relational inference where interactions change over time. We propose  DYnamic multi-Agent Relational Inference (\ours) model, a deep generative model that can reason about \textit{dynamic} relations.  Using a simulated physics system,  we study various dynamic relation scenarios, including periodic and additive dynamics. We  perform comprehensive study on the trade-off between dynamic and inference period, the impact of training scheme, and model architecture on dynamic relational inference accuracy. We also showcase an application of our model to infer coordination and competition patterns from real-world multi-agent basketball trajectories.
\end{abstract}

\section{Introduction}
\label{sec:intro}
Particles, friends, and teams are multi-agent relations at different scales. Learning multi-agent interactions is essential to our understanding of the structures and dynamics underlying many systems. Practical examples include  understanding social dynamics among pedestrians \citep{alahi2016social}, learning communication protocols in traffic \citep{sukhbaatar2016learning,lowe2017multi} and predicting  physical interactions of particles \citep{mrowca2018flexible, li2018learning, sanchez2020learning}. Most existing work on modeling relations  assume  the interactions are  \textit{observed} and train the models with \textit{supervised} learning. For multi-agent trajectories, the interactions are \textit{hidden} and thus  need to be inferred  from data in an \textit{unsupervised} fashion. While one could impose  an interaction graph structure \citep{battaglia2016interaction}, it is difficult to find the  correct structure as the search space is very large \citep{grosse2012exploiting}.  The search task is computationally expensive and the resulting model can potentially suffer from  the model misspecification issue \citep{koopmans1950identification}. 

\begin{wrapfigure}{r}{0.4\textwidth}
\includegraphics[width=\linewidth]{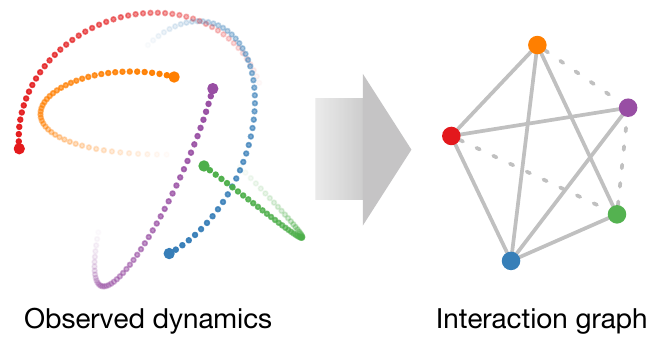}
\caption{Neural Relational Inference for learning the interaction graph. Picture taken from \citep{kipf2018neural}}
\label{fig:nri_particle}
\vspace{-3mm}
\end{wrapfigure}

Relational inference aims to discover hidden interactions from  data and has been studied for decades. Statistical relational learning are based on  probabilistic graphical models such as probabilistic  relational model \citep{kemp2008discovery, getoor2001learning, koller2007introduction, shum2019theory}. However, these methods may require significant feature engineering and high computational costs. Recently, \citet{battaglia2016interaction, santoro2017simple} propose to reason about relations using graph neural networks but still requires supervision.  One exception is Neural Relational Inference (\nri) \citep{kipf2018neural}, a flexible  deep generative model that can infer potential relations in an unsupervised fashion. As shown in Figure \ref{fig:nri_particle}, \nri{} simultaneously learns the dynamics from multi-agent trajectories and    infers their relations. In particular, \nri{} builds upon  variational auto-encoder (VAE) \citep{kingma2013auto} and introduces latent variables to represent  the hidden relations.  Despite its flexibility,  a major limiting factor of \nri{} is that it assumes the relations among the agents are \textit{static}. That is, two agents are either interacting or not interacting regardless of their  states at different time steps, which is rather restrictive.

%


In this paper, we study a more realistic setting: \textit{dynamic}  relational inference.  For example, in game plays, players can coordinate and compete dynamically depending on the strategy.
We propose a novel deep generative model, which we call  DYnamic multi-Agent Relational Inference (\ours{}). \ours{}  encodes  trajectory interactions at different time steps. It  utilizes deep temporal CNN models with pyramid pooling to extract rich representations from the interactions.    \ours{} infers the relations for each sub-sequence dynamically and jointly decode a sequence of relations.

As relational inference is unsupervised, we use simulated dynamics physics systems as  ground truth for validation. We find that the performance of the static  \nri{} model deteriorates significantly with shorter output trajectories, making it unsuitable for dynamic relational inference. In contrast, \ours{} is able to accurate infer the hidden relations with various dynamics scenarios.   We also perform extensive  ablative study to understand the effect of  inference period,  training schemes and model architecture. Finally, We  showcase our \ours{} model on real-world basketball trajectories. 

In summary, our contributions include:
\begin{itemize}
    \item We tackle the challenging problem of  unsupervised learning of hidden dynamic relations given multi-agent trajectories.
    \item We develop a novel deep generative model called \ours{} to handle time-varying interactions and predict a sequence of hidden relations in an end-to-end fashion.
    \item We demonstrate the effectiveness our method on both the simulated physics dynamics and real-world basketball game play datasets. 
\end{itemize}


\section{Related work}
\label{sec:rw}
\paragraph{Deep sequence models} 
Deep sequence models include   both deterministic  models \citep{alahi2016social,li2019enhancing, mittal2020learning} and stochastic  models \citep{chung2015recurrent, fraccaro2016sequential,krishnan2017structured, rangapuram2018deep,chen2018neural,huang2018neural,yoon2019time}. For GAN-like models, \citep{yoon2019time} combine adversarial training and a supervised learning objective for time series forecasting. \cite{liu2019naomi} propose a non-autoregressive model for sequence generation. Compared with GANs, VAE-type models can provide explicit inference and are preferable for our purpose. For instance,  \citet{chung2015recurrent} introduces stochastic layers in recurrent neural networks to model speech and hand-writing. \citet{rangapuram2018deep} parameterizes a linear state-space model for probabilistic time series forecasting. \citet{chen2018neural,huang2018neural} combine normalizing flows with autoregressive models. However, all existing  models only  model the  \textit{temporal} latent states for individual sequences rather than their  \textit{interactions}.
\vspace{-3mm}
\paragraph{Relational inference} Graph neural networks (GNNs) seek to learn  representations over relational data, see several recent surveys on   GNNs and the references therein, e.g. \citep{ wu2019comprehensive,goyal2018graph}. 
Unfortunately, most existing work assume the  graph  structure is \textit{observed} and train with supervised learning.  In contrast,  relational inference aims to discover the \textit{hidden} interactions and  is unsupervised.  Earlier work in relational reasoning \citep{koller2007introduction} use probabilistic graphical models, but requires significant feature engineering. The seminal work of \nri{} \citep{kipf2018neural} use neural networks to reason in dynamic physical systems.  \citet{alet2019neural} reformulates \nri{} as  meta-learning and proposes simulated annealing to  search for graph structures.  Relational inference is also posed as Granger causal inference  for sequences \citep{louizos2017causal, lowe2020amortized}. Nevertheless, all existing work are limited to \textit{static} relations while we focus on \textit{dynamic} relations.

 \vspace{-3mm}
\paragraph{Multi-agent  learning}  Multi-agent  trajectories   arises frequently  in reinforcement learning (RL) and imitation learning (IL) \citep{albrecht2018autonomous, jaderberg2019human}. Modeling agent interactions given  dynamic observations  from the environment remains  a central topic.  In the RL setting,  for example, \citet{sukhbaatar2016learning} models the control policy in a fully cooperative multi-agent setting and applies  a GNN to represent the communications.   \citet{le2017coordinated}  models the agents coordination as a latent variable  for imitation learning.  \citet{song2018multi} generalizes GAIL \citep{ho2016generative} to multi-agent through a shared generator. However, these coordination models only capture the global interactions implicitly without the explicit graph structure.  \citet{tacchetti2018relational} combines GNN with a forward dynamics model to model multi-agent coordination but also requires supervision.  \citet{grover2018learning} directly models the episodes of interaction data   with GNs for learning  multi-agent policies.  Our method instantiates the multi-agent imitation learning framework, but focuses on relational inference. Our approach is also applicable to dynamic modeling in model-based RL. 

\section{Dynamic Multi-Agent Relational Inference}
\label{sec:method}
Given a collection of multi-agent trajectories, we aim to reason about their hidden  relations over time. First we describe the underlying probabilistic inference problem. 



\subsection{Probabilistic inference formulation}
For each agent $i \in \{1, \cdots, N \}$, define its state (coordinates) as $x_t \in \mathbb{R}^D$.  A trajectory $\tau^{(i)} =(x_1, x_2,\cdots, x_T) $ is  a sequence of states  that are sampled from  a  policy.    Given   trajectories from $N$ agents $\{\tau^{(i)}\}_{i=1}^N$, dynamic relational inference aims to infer the pairwise interactions of $N$ agents at every time step.  Mathematically speaking, the joint distribution of the trajectories can be written as:
\begin{wrapfigure}{r}{0.38\textwidth}
\includegraphics[width=1.0\linewidth]{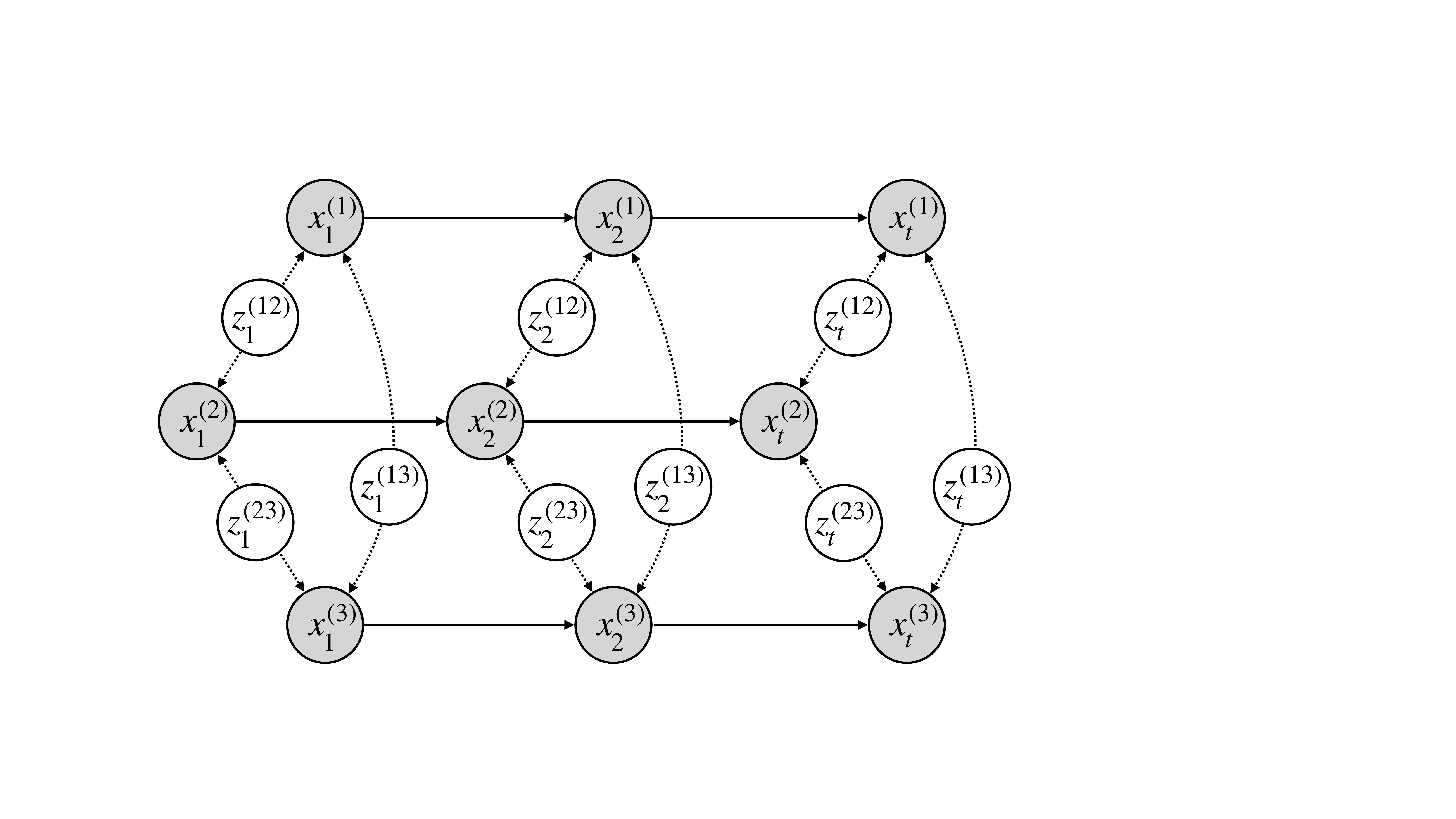}
\caption{Probabilistic graphical model representation of dynamic multi-agent relational inference. Shaded nodes are observed and unshaded nodes are hidden variables.}
\label{fig:pgm}
\vspace{-8mm}
\end{wrapfigure}
\begin{eqnarray}
    p(\tau^{(1)},\cdots, \tau^{(N)}) = \prod_{t=1}^T  p(\V{x}_{t+1}| \V{x}_t, \cdots, \V{x}_1)
\end{eqnarray}
where $p(\V{x}_{t+1}|\V{x}_t, \cdots, \V{x}_1)$ represents the state transition dynamics. We use the bold form $\V{x}_t := (x_t^{(1)}, \cdots,x_{t}^{(N)})$  to indicate the concatenation of all agents observations and $\V{x}_{<t}:=(\V{x}_1, \cdots, \V{x}_t)$.

We introduce latent 
variables $z_t^{(ij)}$ to denote the interactions between agent $i$ and $j$ at time $t$. To make the problem tractable, we restrict $z_t^{(ij)} $ to be categorical, representing discrete interactions such as coordination or competition. We assume that the dynamics model can be decomposed into  the individual dynamics, in conjunction with the pairwise interaction.  This substantially reduces the dimensionality of the distribution and simplifies learning.   Therefore, we  can rewrite the transition dynamics  as:
\begin{eqnarray}
  p(\V{x}_{t+1}| \V{x}_{<t}) \approx \int_{\V{z}} \prod_{i=1}^N p(x_{t+1}^{(i)}|x_{<t}^{(i)},z_t^{(ij)})  \prod_{i=1}^N \prod_{j=1, j\neq i}^N p(z_t^{(ij)}|x_{<t}^{(i)}, x_{<t}^{(j)}) d\V{z}
\end{eqnarray}
Here  each  $p(x_{t+1}^{(i)}|x_{<t}^{(i)})$ captures the state transition dynamics of a single agent.  $p(z_t^{(ij)}|x_{<t}^{(i)}, x_{<t}^{(j)})$ represents the  latent interactions between two agents. Figure \ref{fig:pgm} visualizes the graphical model representation for three agents over $t$ number of time steps. The shaded nodes represent observed variables and the unshaded nodes are latent variables.  Dynamic relational inference is to estimate distributions of the hidden variables $\{z_t^{(ij)}\}$ at different time steps.

\subsection{Dynamic Multiagent Relational Inference (\ours{})}
\begin{figure}[t!]
        \centering
        \includegraphics[trim=10 50 10 100,width=0.9\linewidth]{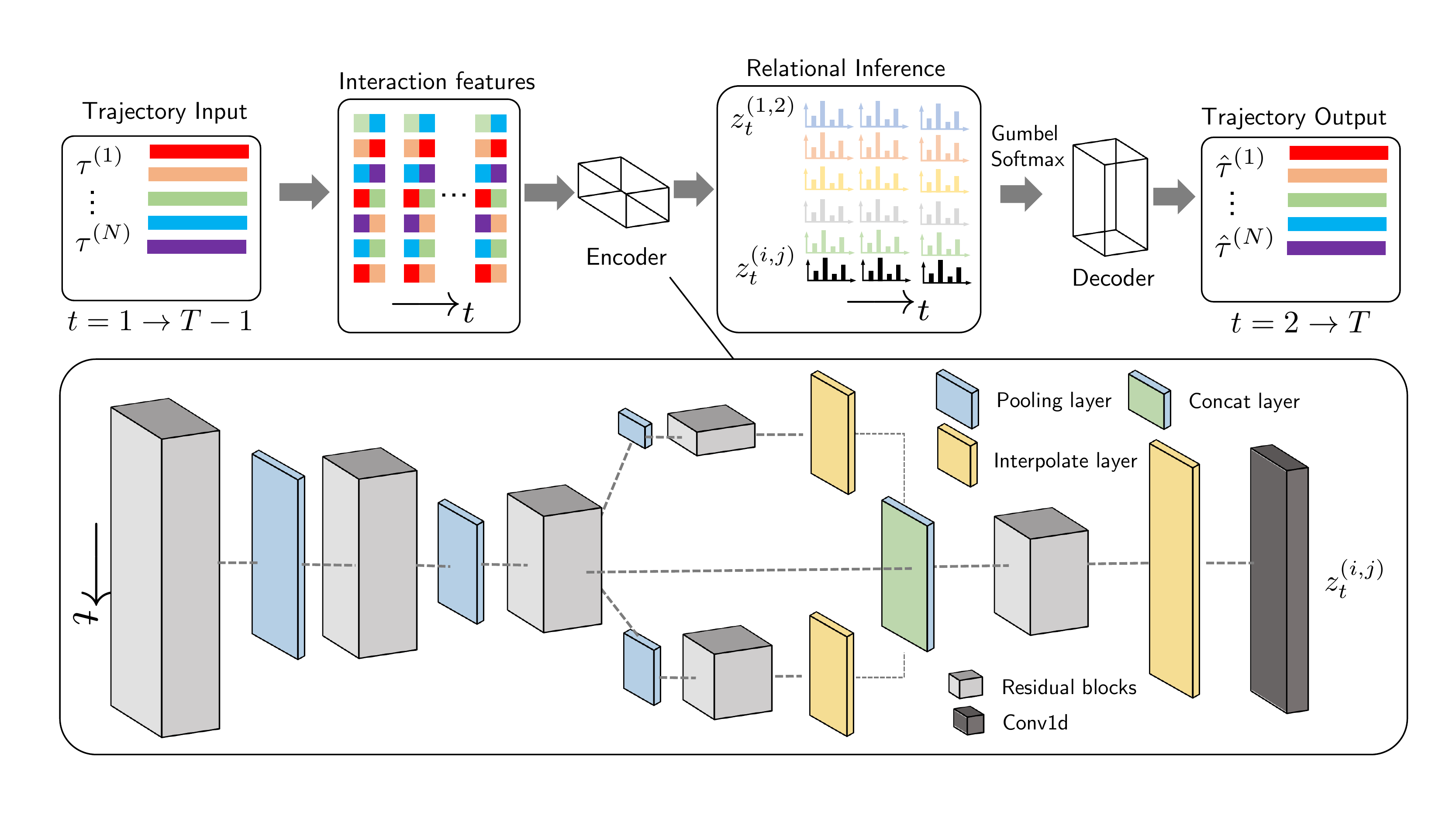}
        \caption{Visualization of the \ours{} model. It infers pairwise relations at different time steps given trajectories. 
        The bottom diagram shows the  details of the encoder and the decoder is the same as \nri{}.
        }
        \label{fig:modelarc}
        \vspace{-2mm}
\end{figure}

%
We propose a deep generative model: Dynamic multi-Agent Relational Inference (\ours). 
Given the trajectories $(\V{x}_1, \cdots, \V{x}_T)$ of all agents,  \ours{} first concatenates the trajectories based on a fully connected graph. The concatenated trajectories are used as  interaction features for the encoder. Then we sample the  sequence of relations from the encoded hidden states. Finally, we generate the future trajectory predictions conditioned on the sampled relations.   Figure~\ref{fig:modelarc} visualizes the overall architecture of our model which encodes and decodes multi-agent trajectories.  The bottom cut-out diagram shows the architecture of our  encoder. 

\paragraph{Encoder.}   A key ingredient of \ours{} is an encoder that is inspired by PSPNet \citep{zhao2017pyramid} to learn rich representations of trajectories at different scales.  In particular, we define a residual block as a two-layer CNN with residual connections~\citep{he2016deep}. Our  encoder has four modules:  feature extraction, pyramid pooling, an aggregation module, and an interpolation module.  
\begin{itemize}[leftmargin=*]
\item \textit{Feature extraction}: the feature extraction module consists of  multiple  residual blocks  interleaved with pooling layers to extract rich temporal features. 
\item \textit{Pyramid pooling}: the pyramid pooling module learns multi-scale temporal representations from the extracted features. First, the output of the feature extraction module is downsampled by 2x and 5x through average pooling. Then, the downsampled features are passed through two residual blocks and finally upsampled  by 2x and 5x  to generate features which are of the same size as the input. The representations learned at 2x and 5x resolutions are concatenated with the input to generate composite multi-scale features. 
\item \textit{Aggregation module}:  a 1-D convolution  that aggregates the multi-scale features.
\item \textit{Interpolation module}: it  average-pools the aggregated features corresponding to the dynamic period.  Then the outputs are upsampled through nearest neighbours interpolation to obtain the  hidden presentations for the relations.
\end{itemize}
%



\paragraph{Sampling.} We utilize  variational inference  \citep{kingma2013auto} to sample the latent variables from hidden representations. Specifically,  assume the interaction posterior $z_t^{(ij)}$ to be categorical:
\[q_\phi(z_t^{(ij)}|\V{x}_{<t}) \sim \text{Cat}(p_1,\cdots, p_k) \]
Using the Gumbel-Max trick \citep{jang2017categorical}, we  can reparameterize the  categorical distribution as:  $z_t^{(ij)} = \texttt{Softmax}(h_t^{(ij)} + g_t^{(ij)} )$. Here  $h_t^{(ij)}$ is the hidden states of the encoder  and   $g_t^{(ij)}$ is  a random Gumbel vector.  
Note that a defining feature of  \ours{} is that the  latent variable $z_t^{(ij)}$ is time-dependent, requiring fine-grained modeling. Our  encoder  ensures that the learned representations are expressive enough to capture such complex dynamics.



\paragraph{Decoder.}
Given  the sampled  latent variables,  the decoder generates the prediction auto-regressively following a Gaussian distribution:
\begin{eqnarray}
 & p(x_{t+1}^{(i)}|\V{x}_{<t},z^{(ij)}_{t}) = \T{N}(x_{t+1}^{(i)}| \mu^{(i)}_{t+1}, \sigma^2 I)  \\
 & \mu_{t+1}^{i} = f_{\text{dec}}( \sum_{j\neq i}\sum_k z_{t,k}^{(ij)} u_k; \theta), \quad u_k = f_{\text{mlp}}^k( x_{t}^{(i)}, x_{t}^{(j)} )
 \label{eqn:decoder}
\end{eqnarray}
Here the output $x_{t+1}^{(i)}$ is reparameterized by a Gaussian distribution with mean  $\mu_{t+1}^{(i)}$ and a fixed standard deviation $\sigma^2$. The mean vector $\mu^{(i)}_{t+1}$ of agent $i$ is computed by  aggregating the hidden states of all other agents. 
%
%
We use a separate MLP to encode the previous inputs into different type of  edges in a k-dimensional one-hot vector $z_t^{(ij)}$. 
To generate long-term predictions using the  model in Eqn. \eqref{eqn:decoder}, we can also incorporate the predictions from the  previous time step.
The decoder architecture is the same as  in \nri{} at a given time step, which consists of message passing GNN operations, followed by a GRU \citep{cho2014learning} decoder.

\paragraph{Inference.}
At every time step $t$, we learn a different distribution for the hidden relation  $z^{(ij)}_t$. We assume a uniform prior for $p_\theta(\V{z}_{t})$ and use ELBO as the optimization objective:
\begin{eqnarray}
 \T{L}_{\text{ELBO}} &=& \mathbb{E}[\log p_\theta(\V{x}_{<T}|\V{z}_{<T})] - \beta d_{\text{KL}}[q_\phi(\V{z}_{<T}|\V{x}_{<T})||p_\theta(\V{z}_{<T})]]  \\ \nonumber
  &=& -\sum_{i=1 }^N\sum_{t=1}^{T} \frac{({\mu}_{t}^{(i)} - x_{t}^{(i)})^2}{2\sigma^2} + \beta\sum_{i,j}^N\sum_{t=1}^{T} H(q_\phi(z_{t}^{(ij)}|\V{x}_t)) 
 \label{eqn:elbo}
\end{eqnarray}
where the mean vector ${\mu}_t^{(i)}$ is parameterized by the decoder. $H$ is the entropy function and $\beta$ balances the two terms in ELBO \citep{higgins2016beta}. 




\section{Experiments}
\label{sec:exp}
We conduct extensive experiments  on simulated physics dynamics and real-world basketball trajectories. The majority of our experiments are based on the physics simulation in the Spring environment. This is ideal for model verification and ablative study as  we know the ground truth relations. 
\subsection{Physics Simulations}
\begin{wrapfigure}{R}{0.35\linewidth}
\includegraphics[width=\linewidth]{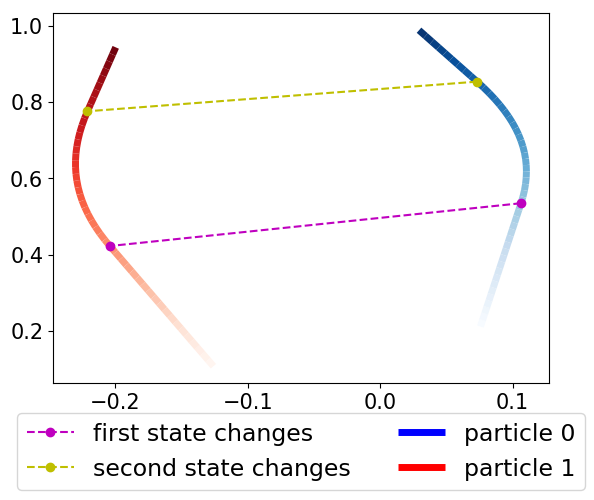}
\caption{Example trajectories of two particles. with two relation changes. The trajectories start from the end with lighter color and gradually become darker. }
\label{fig:demo}
\vspace{-3mm}
\end{wrapfigure}

\paragraph{Data Generation}
The Spring environment \citep{kipf2018neural} simulates the movements of a group of particles connected by a spring in a box. The hidden relation is whether there is a spring connecting the two particles. To simulate dynamic relations, we generate the trajectories by removing and adding back the springs following certain  patterns. Figure ~\ref{fig:demo} visualizes the trajectories resulting from such dynamic relations. Starting from the bottom, the two-particle trajectories appear as straight lines and bend in the middle due to the spring force, and return to straight lines after the removal of the spring. 

We define the number of time steps between the change of relations  as the \textit{dynamic period}. The primary challenges for dynamic relation inference arise along two dimensions:
\begin{enumerate}[wide, labelwidth=!, labelindent=0pt]
\item The shorter the dynamic period, the more frequent the relation changes.  Hence, it becomes more difficult to infer relations with shorter dynamic period.
\item If the dynamic period itself also changes, then the task becomes much harder because the model  also needs to adapt to the unknown period in the changing relations. Note that the way  relations change in the trajectories must follow certain pattern and not be completely random.  Otherwise it would be impossible to learn anything meaningful.
\end{enumerate}
We experiment with two types of dynamic relations: periodic dynamics   and  additive dynamics. For periodic dynamics, we generate the trajectories by periodically removing and adding back the springs. We investigate the model performance by generating data with different frequencies of periodicity. For additive dynamics, we assume the dynamic period is increasing arithmetically. Each trajectory is of length $50$ and the decoding   length is $40$, see  details of the generated dataset in Appendix.

%

\vspace{-3mm}

\paragraph{Baselines and Setup}
We consider several baselines for comparison: (1) \nri{} (static): unsupervised  \nri{}  with an encoder  trained using the entire trajectory and infer repeatedly over time. This corresponds to \nri{} (learned) in \citep{kipf2018neural}.
(2)  \nri{} (adaptive):  \nri{} (static) with an encoder trained over sub-trajectories. The encoding length corresponds exactly to the dynamic period of the dataset. We use the \nri{} (static)  decoder  to predict the entire trajectory in an auto-regressive fashion.
(3) Interaction Networks (\inter{}) \citep{battaglia2016interaction}: a supervised GNN model which uses the ground truth relations  to  predict  future trajectories. We include this supervised learning model as the ``gold standard'' for our inference tasks. 



In practice, we do not know the dynamic period  beforehand. Therefore, how often we infer the relations is a difficult choice: rare inference would miss the time steps where relations change while predicting too frequently  introduces more latent variables and complicates the inference. To investigate this trade-off, we  define  \textit{inference period} as the number of time steps between two predicted relations. Unless otherwise noted,  the inference period in our experiments is the same as the dynamic period.
All the models are trained to predict the sequence in an auto-regressive fashion: the prediction of the current time step is fed as the input to the next time step. We use Adam~\citep{kingma2014adam} optimizer with learning rate $5e^{-4}$ and weight decay $1e^{-4}$ and train for 300 epochs.



\subsubsection{Pathological Cases of Neural relational inference}


It is known that latent variable models suffer from the problem of identifiability \citep{koopmans1950identification}, which means certain parameters, in principle, cannot be estimated consistently.  \nri{} infers correlation-like relations between trajectories which highly depend on the length of the time lag.  To test this hypothesis,  we  follow  the exact same setting as \citet{kipf2018neural} to  infer the interaction graph. Instead of decoding $50$ time steps,   we  vary the   length of input and output sequence.  




Table~\ref{tab:nri_exp} summarizes the inference accuracy with different sequence length in the encoder and decoder.  We can see that the performance of \nri{} deteriorates drastically with shorter training sequences, simply increasing the capacity of the encoder (\nri{}++) does not help. One plausible explanation is that  \nri{} is learning correlation-like interactions. Shorter decoding sequences carry less information about  correlations, making it harder to learn. Meanwhile, we also observed that using auto-regressive can achieve better inference accuracy compared to teacher forcing. 
The pathological cases  highlight the issue of  \nri{} for dynamic relational inference. If the interactions  change  frequently every few time steps, repeatedly applying \nri{} to different time steps would suffer from short decoding sequences.  Therefore, having a model that can jointly infer a sequence of relations  is critical.

\begin{table}[h]
\centering
\caption{Inference accuracy (\%) of \nri{} trained with trajectory lengths. Note that the performance deteriorates significantly when the output length decreases. For \nri{}++,  we added two more hidden layers to the  MLP encoder of \nri{}.}
\begin{tabular}{c|cccc | cccc}
\toprule
& \multicolumn{4}{c|}{Teacher Forcing} & \multicolumn{4}{c}{Auto-regressive} \\ 
 Output Length & 40        & 20        & 8         & 4     & 40        & 20        & 8         & 4     \\
\midrule
  \nri{}                                                                   & 0.99      & 0.65      & 0.63      & 0.54   & 0.99      & 0.81      & 0.80      & 0.69      \\\
 \nri{}++  & 0.99      & 0.66      & 0.63      & 0.53     & 0.99      & 0.80      & 0.80      & 0.70\\ 
\bottomrule
\end{tabular}
\label{tab:nri_exp}
\vspace{-3mm}
\end{table}

\subsubsection{Dynamic Relational Inference Comparison}
We compare the performance of different models for dynamic relational inference tasks. 
\paragraph{Periodic dynamics}  In the periodic scenario, the dynamic period is fixed. We generate four datasets with a dynamic period of $40$, $20$, $8$, $4$ to simulate  relational dynamics  with increasing frequency. Table.~\ref{tab:encoder_comp} columns ``40, 20, 8, 4''  show  the  trajectory prediction mean square error (MSE) and interaction inference accuracy comparison of different methods. 

We can see that all methods can achieve almost perfect predictions of the trajectories with very low MSE. However, there is a sharp difference in relational inference accuracy. \nri{} (static) is incapable of learning dynamic interactions. \nri{}(adaptive) can learn but has lower accuracy due to short decoding sequences.  With a more expressive encoder and joint decoding,  \ours{} is able to reach higher accuracy.  When the  dynamic period is  very small at 4, even \ours{} struggles slightly, suggesting the fundamental difficulty  with frequently changing dynamics.

\begin{table}[htbp]
\centering
\caption{Performance comparison  for ours and the baselines in both the periodic (40,20,8,4) and additive (Add) dynamic scenarios.  MSE is for  trajectory prediction   and Accuracy quantifies the dynamic relational inference performance. }
\label{tab:encoder_comp}
\begin{tabular}{c|cccc|c|| cccc |c}
\toprule
 \textbf{Dynamic}& \multicolumn{5}{c||}{MSE $\downarrow$} & \multicolumn{5}{c}{Accuracy $\uparrow$}  \\ 
\textbf{Period} & 40 & 20 & 8 & 4 & Add & 40 &   20 & 8  & 4 & Add \\
\midrule
\nri{} (static)     & 2.2e-4 & 5.2e-3 & 2.7e-3 & 2.4e-3 & 3.6e-3  & 0.99 & 0.52 & 0.51 &  0.50 &0.53 \\
\nri{} (adaptive)     & 2.2e-4 & 2.7e-3 & 1.3e-3 & 5.9e-4 & 3.1e-3 & 0.99 & 0.81 & 0.80 &  0.69 &0.81 \\
\ours{}& \textbf{2.6e-5} & \textbf{4.1e-5} & \textbf{4.6e-6} & \textbf{3.6e-6} &\textbf{7.6e-6} & \textbf{0.99} & \textbf{0.92} & \textbf{0.91} & \textbf{0.74} & \textbf{0.87}\\
\midrule
\inter{} & 2.9e-5 & 2.3e-5 & 4.3e-5 & 4.7e-5 &3.9e-5  & 0.99 & 0.99 & 0.99 & 0.98 & 0.99\\ 
\bottomrule
\end{tabular}
\vspace{-3mm}
\end{table}

\paragraph{Additive Dynamics}
In the additive scenario, we allow the dynamic period itself to increase arithmetically. We increase the dynamic period in steps of 4 starting from a dynamic period of 4 timesteps. In a sequence of 40 timesteps, this implies that the relations (spring connection) get flipped at timesteps 4, 12 and 24.   We use four \nri{}(static) models, each trained  separately with  4, 8, 12 and 16 encoding timesteps. We  combine the ensemble model predictions into \nri{}(adaptive).  Table.~\ref{tab:encoder_comp} ``Add'' columns show the  performance comparison. Similar to the periodic scenario, \ours{} outperforms the baselines in this challenging task as well. Note that \nri{}(adaptive) is a close competitor w.r.t inference accuracy, but it is a four model ensemble and takes a long time to train.



\subsubsection{Ablative Study}
We perform ablative study to further validate our experiment design and understand the  behavior of \ours{}. In particular, we study the trade-off between dynamic and inference period, the effect of training scheme, as well as the ablative study of model architecture design.

\begin{wraptable}{rb}{6cm}
\centering
\caption{Inference accuracy for different combinations of  dynamic and inference periods with \ours{}.}
\label{tab:trade-off}
\begin{tabular}{c|c|cccc}
\toprule
\multicolumn{2}{c|}{\textbf{Dynamic Period}} & 40 & 8 & 4 \\ \hline
\multirow{3}{*}{\begin{tabular}[c]{@{}c@{}}\textbf{Inference} \\ \textbf{Period} \end{tabular}}
 & 40   & \textbf{0.99}  & 0.50 & 0.50\\ 
 & 8         & 0.88  & \textbf{0.80} & 0.50 \\ 
 & 4          & 0.80 & 0.76 & \textbf{0.74}  \\
\bottomrule
\end{tabular}
\vspace{-6mm}
\end{wraptable}


\paragraph{Dynamic vs. Inference Period.} 
To understand the relations between dynamic  and  inference period, we repeat the periodic scenario experiments by varying both  dynamic and inference period in  $40, 8, 4$ time steps.

In Table~\ref{tab:trade-off}, we observe that dynamic relational inference reaches the highest accuracy when the inference period matches the dynamic period. If the inference period is longer than the dynamic period,  the model can miss the changes in the relations and completely fails to perform inference. Meanwhile, if the inference period is shorter  than the dynamic period, the model still can learn  but suffers from low accuracy. This is potentially due to the extra uncertainty introduced by estimating more latent variables.


\paragraph{Decoder Training Scheme}
\begin{figure}[t!]
        \centering
        \includegraphics[width=0.95\linewidth]{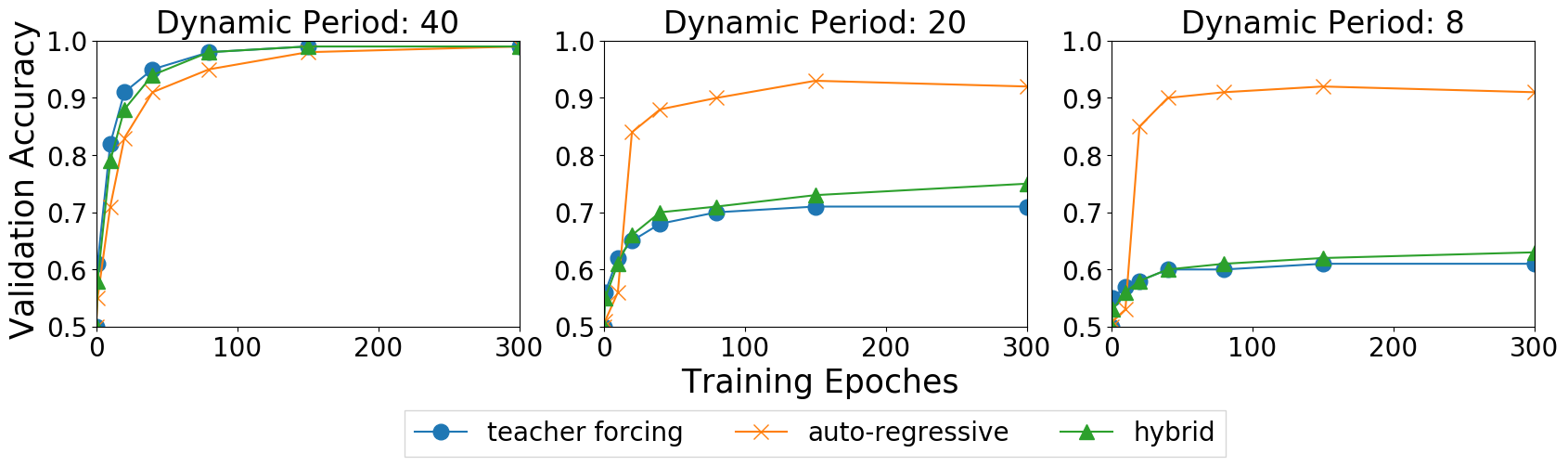}
        \caption{Learning curve of \ours{} trained with teacher forcing (blue), auto-regressive (yellow), and hybrid (green) in decoder. Hybrid model is  first trained with teacher forcing in the beginning 30 time steps and then auto-regressively in the later 10 time steps. We report the relational inference accuracy on the validation data for different dynamic periods.}
        \label{fig:compare}
        \vspace{-2mm}
\end{figure}
Another fundamental challenge in sequence prediction is  covariate shift \citep{bickel2009discriminative} -- a mismatch between  distribution  in training and testing -- due to sequential dependency. Common solutions to mitigate covariate shift include  teacher forcing \citep{williams1989learning} and scheduled sampling \citep{bengio2015scheduled}.  However, all these work are focused the prediction of \textit{observed} sequence while our  sequence predictions are on the \textit{latent} variables. It is not evident that covariate shift exists in this setting. We demonstrate the empirical evidence for the effect of different training schemes on the accuracy of relational inference.

Quite surprisingly, we found that auto-regressive training is most effective for dynamic relations inference. Figure.~\ref{fig:compare} summarizes the difference in learning curve between using teacher forcing and auto-regressive for different dynamic periods. We also include a version of scheduled sampling (hybrid): in the first 30 time-steps, we  train the model  with teacher forcing and then switch to auto-regressive in the last 10 time-steps. We observe that while teacher forcing converges faster, it leads to lower accuracy. This observation is consistent across different dynamic periods. Therefore,   auto-regressive training is preferred for dynamic relation inference.

\subsection{Real-World Basketball Data Experiments}
To showcase the practical value of dynamic relational inference, we apply \ours{} to a real-world basketball trajectory dataset. The goal of the experiment is to extract meaningful ``hidden'' relations in competitive basketball plays. 
The basketball dataset contains trajectories for 10 players in a game. As the ground-truth relations are unknown, we use the trajectory prediction MSE and negative ELBO as  in-direct measures for the dynamic relational inference performance. We assume there are two types of hidden relations:  coordination and competition. We defer the details of the dataset and training setup to the Appendix.

\begin{figure}[t!]
    \centering
    \begin{minipage}[c]{0.19\textwidth}
    \includegraphics[width=\textwidth]{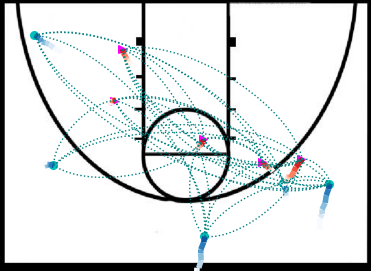}
    \centering
    \end{minipage} 
    \hfill 	
    \begin{minipage}[c]{0.19\textwidth}
    \includegraphics[width=\textwidth]{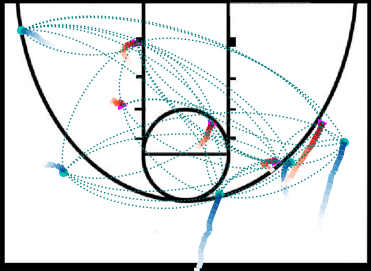}
    \centering
    \end{minipage} 
    \hfill 	
    \begin{minipage}[c]{0.19\textwidth}
    \centering
    \includegraphics[width=\textwidth]{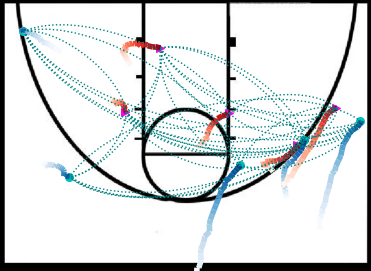}
    \end{minipage}
    \begin{minipage}[c]{0.19\textwidth}
    \includegraphics[width=\textwidth]{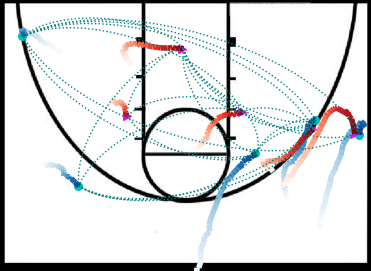}
    \centering
    \end{minipage}
    \begin{minipage}[c]{0.19\textwidth}
    \includegraphics[width=\textwidth]{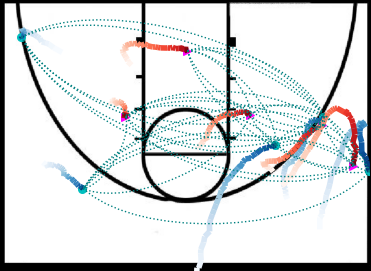}
    \centering
    \end{minipage}
    \hfill
    \begin{minipage}[c]{0.19\textwidth}
    \centering
    \includegraphics[width=\textwidth]{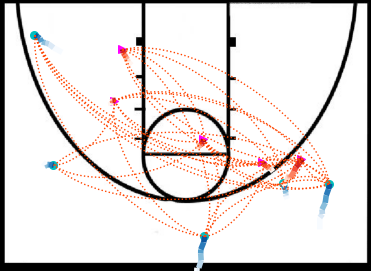}
    \end{minipage} 
    \hfill 	
    \begin{minipage}[c]{0.19\textwidth}
    \centering
    \includegraphics[width=\textwidth]{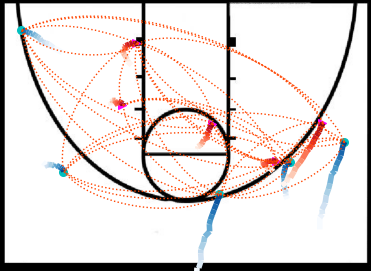}
    \end{minipage} 
    \hfill 	
    \begin{minipage}[c]{0.19\textwidth}
    \centering
    \includegraphics[width=\textwidth]{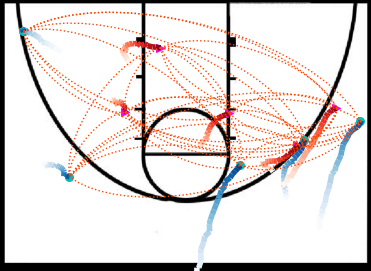}
    \end{minipage}
    \hfill
    \begin{minipage}[c]{0.19\textwidth}
    \centering
    \includegraphics[width=\textwidth]{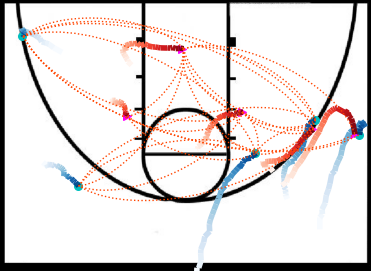}
    \end{minipage}
    \begin{minipage}[c]{0.19\textwidth}
    \centering
    \includegraphics[width=\textwidth]{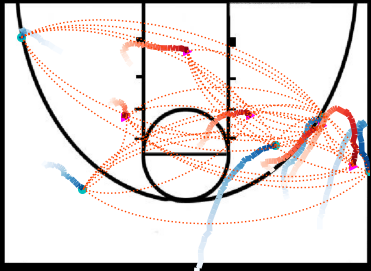}
    \end{minipage}
    \vspace{-2mm}
    \caption{Visualization of the inferred relations (dashed links) in the basketball players trajectories over time by \ours{} with an inference period of 8. The blue dashed links in the  top  are the inferred  interactions from the same team (coordination) and red dashed links in the bottom are  from different teams (competition). Different columns represent different time steps.}
    \label{fig:main_basketball}
\end{figure}

We report performance comparisons for different inference periods. As shown in Table~\ref{tab:basketball}, we observe lower MSE loss and negative ELBO with shorter inference period. Intuitively, the interactions in the real world may change constantly, thus shorter inference period can capture the dynamics better.  \ours{} outperforms the baselines  in trajectory prediction MSE and negative ELBO loss. Notice that  \nri{}(adaptive) is using encoder and decoder that are trained separately and this results in a high negative ELBO loss on the test set. Fig.~\ref{fig:main_basketball} visualizes a sample trajectory of 10 basketball players with inferred relations from \ours{}  over different time steps. We separate coordination and competition interactions in different rows. In Fig.~\ref{fig:main_basketball},  Kobe Bryant is moving along with three-point line and guarded by a defender. We can see clear attention drawn to the specific players throughout the play. See  Appendix for other inferred relations. 


\begin{table}[t]
\centering
\centering
\begin{tabular}{c|cccc|cccc}
\toprule
 & \multicolumn{4}{c}{MSE $\downarrow$} & \multicolumn{4}{c}{NELBO $\downarrow$} \\ 
\textbf{Inference Period} & 40 & 20 & 8 & 4 & 40 & 20 & 8 & 4 \\ \hline
\nri{}(static) & 2.3e-3 & - & - & -& 13.71 & - & - & -\\
\nri{}(adaptive) & 2.3e-3 & 3.0e-2 & 3.3e-2 & 9.7e-3 & 13.71 & 303.10 & 337.54 & 96.76\\
\ours{} & 2.2e-3 & 8.4e-4 & 4.6e-4 &  1.8e-4 & 12.65 & 6.16 & 4.38 & 3.67\\
\bottomrule
\end{tabular}
\caption{Performance comparison for \ours{} and baselines on the real-world basketball trajectory dataset with different inference periods $40$, $20$, $8$ and $4$. }
\label{tab:basketball}
\end{table}

\section{Conclusion}
\label{sec:con}
We investigate unsupervised learning of dynamic relations  in multi-agent trajectories.  We propose a novel deep generative  model: Dynamic multi-Agent Relational Inference (\ours) to infer changing relations over time. We conduct extensive experiments using a simulated physics system to study the performance of   \ours{} in handling various dynamic relations. We perform ablative study to understand the effect of dynamic and inference period, training scheme and model design choice.  Compared with static \nri{} and its variant, our \ours{} model  demonstrates significant improvement  in simulated physics systems as well as in a real-world basketball trajectory dataset. 


\newpage

\bibliographystyle{iclr2021_conference}
\bibliography{ref}
\clearpage

\newpage

\appendix

\section{Model Implementation Details}
In this section, we include some details about the model implementation, especially the encoder part. Our encoder is analogical to ResNet CNNs~\citep{he2016deep} used in the field of image recognition, where the task can be abstracted to be a classification problem on 1D dimension. Meanwhile, inspired by PSPNet used in visual scene semantic parsing ~\citep{zhao2017pyramid}, we add additional 2 global feature extractors to combine the whole-sequence (global) features  and the sub-sequence (local) features.

for ~\ours{}, each residual block shown in Fig.~\ref{fig:modelarc} consists of 4 skip connections structure.

\begin{figure}[h]
\includegraphics[width=0.8\linewidth]{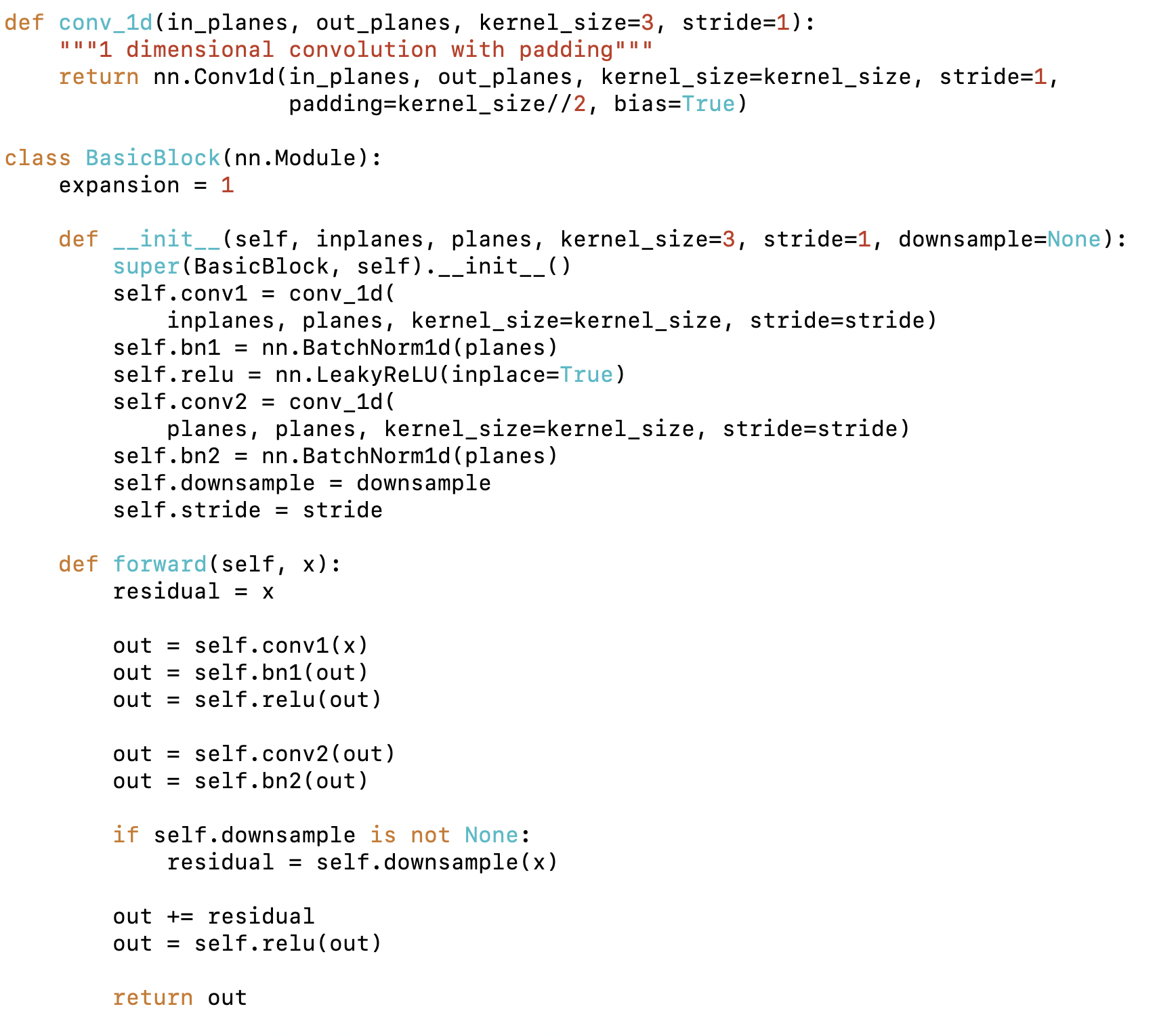}
\label{fig:code_snippet_basic_block}
\caption{Pytorch code snippet of the Residual Block used in \ours{} encoder.}
\end{figure}

\begin{figure}[t!]
    \centering
     \begin{minipage}[c]{0.19\textwidth}
    \includegraphics[width=\textwidth]{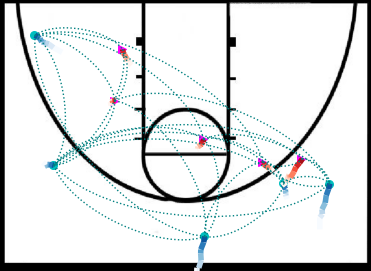}
    \centering
    \end{minipage} 
    \hfill 	
    \begin{minipage}[c]{0.19\textwidth}
    \includegraphics[width=\textwidth]{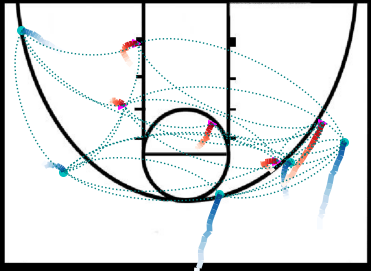}
    \centering
    \end{minipage} 
    \hfill 	
    \begin{minipage}[c]{0.19\textwidth}
    \centering
    \includegraphics[width=\textwidth]{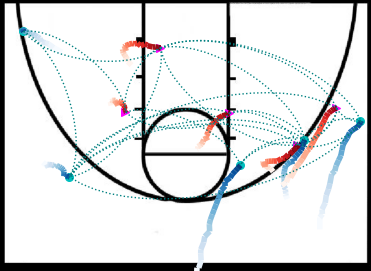}
    \end{minipage}
    \begin{minipage}[c]{0.19\textwidth}
    \includegraphics[width=\textwidth]{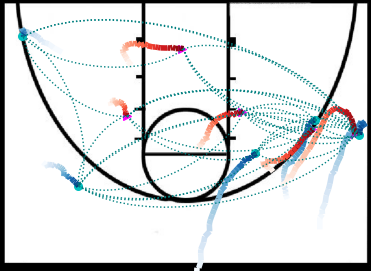}
    \centering
    \end{minipage}
    \begin{minipage}[c]{0.19\textwidth}
    \includegraphics[width=\textwidth]{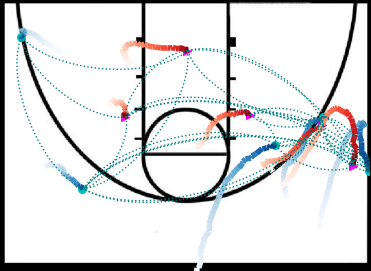}
    \centering
    \end{minipage}
    \hfill
    \begin{minipage}[c]{0.19\textwidth}
    \centering
    \includegraphics[width=\textwidth]{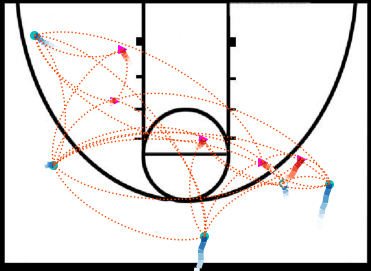}
    \end{minipage} 
    \hfill 	
    \begin{minipage}[c]{0.19\textwidth}
    \centering
    \includegraphics[width=\textwidth]{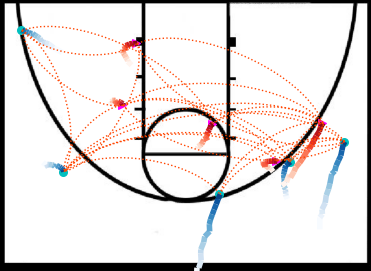}
    \end{minipage} 
    \hfill 	
    \begin{minipage}[c]{0.19\textwidth}
    \centering
    \includegraphics[width=\textwidth]{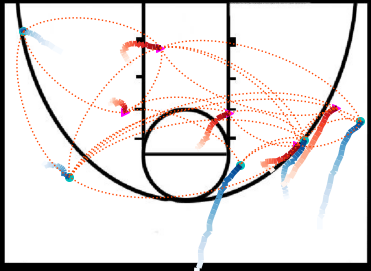}
    \end{minipage}
    \hfill
    \begin{minipage}[c]{0.19\textwidth}
    \centering
    \includegraphics[width=\textwidth]{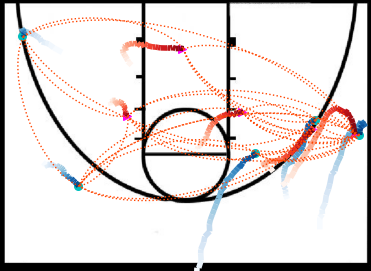}
    \end{minipage}
    \begin{minipage}[c]{0.19\textwidth}
    \centering
    \includegraphics[width=\textwidth]{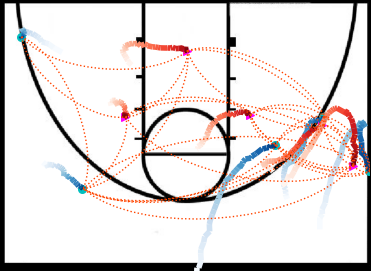}
    \end{minipage}
    \caption{Visualization of the basketball players trajectories with inference period = 8. The top  row visualizes the inferred interactions from the same team (coordination) and the bottom row visualizes the inferred interactions from different teams (competition). Different columns represent different time steps. }
    \label{fig:app_basketball}
\end{figure}

\section{Experimental Details}
\paragraph{Particle dataset}
In general, we use the same pre-processing in \nri{}. Each raw simulated trajectory has length of 5000 and we sample with frequency of 100 so that each sample has length of 50 in our dataset. Correspondingly, the value of dynamic period/inference period matches the length of sample in our dataset. For instance, dynamic period = $10$ means that the in the raw trajectory, the state of a node changes every 1000 time steps. In addition, The value of trajectories are all normalized to range of $[0,1]$ and the evaluation is done on the same range as well.

\paragraph{Basketball dataset details} The basketball dataset consists of trajectory from 30 teams. The raw trajectory is captured with frequency of 25 ms. For our experiment, we sample the trajectory with frequency of 50 ms for more evident player movements. We  use a inference period that matches the length of sample. For instance, inference period = $10$ means that our model produce prediction every 500 ms.  The resulting dataset include 50,000 training samples, 10,000 validations samples and 10,000 test samples.

We normalize the values of the trajectories  to range [0,1]   and train all the models in an auto-regressive fashion. We use the same training set up as in  physics simulation experiments with a batch size of $64$. 

\section{Additional Experiments}
\paragraph{Stochastic dynamics}
In order to make the problem even harder and to unify all the previous settings, we generate a dataset where the edge types are flipped randomly with a probability $p$ after each dynamic period of 4 timesteps. The static data generation corresponds to $p=0$ and the periodic dynamics corresponds to $p=1$. Table~\ref{tab:app_encoder_comp} shows the MSE and inference accuracy of \nri{}, \ours{} and Interaction Networks on the stochastic dataset for flipping probabilities $p=0.8$ and $p=0.9$.


\begin{table}[htbp]
\centering
\caption{Qualitative results for stochastic dynamics. Accuracy improves by increasing the model capacity. In the training, The inference period of the two \ours{} match with the dynamic period.}
\label{tab:app_encoder_comp}
\begin{tabular}{c|cc|cc}
\toprule
 & \multicolumn{2}{c}{MSE} & \multicolumn{2}{c}{Accuracy} \\ 
\textbf{Flipping Probability} & 0.8 & 0.9 & 0.8 & 0.9 \\
\midrule
\nri{}    & 1.4e-3 & 2.3e-3 &  0.59 & 0.60 \\
\ours{} & 8.3e-4 & 1.8e-3 & 0.57 & 0.63 \\
\midrule
\inter{} (Supervised)  & 4.5e-5 & 4.2e-5 & 0.99 & 0.99\\
\bottomrule
\end{tabular}
\end{table}

\begin{table}[h]
\centering
\caption{Results with and without average pooling in the interpolation module of \ours{}.}
\label{tab:app_average_ablation}
\begin{tabular}{c|c|c}
\toprule
 & MSE & Accuracy \\ 
 \midrule
\ours{} without average pooling  & 1.8e-5 & 0.59 \\
\ours{} with average pooling & 4.1e-5 & 0.92 \\
\bottomrule
\end{tabular}
\end{table}
\paragraph{The Effect of Average Pooling}
We perform an ablation study where we remove the average pooling corresponding to the inference period in the interpolation module to study the effect of this average pooling on the results. We find that without average pooling, inference accuracy decreases from 0.92 to 0.59 (as shown in Table~\ref{tab:app_average_ablation}) for inference period at 20. Here we drop the average pooling corresponding to the inference periods and directly interpolate to the sequence length.  This shows that intermediate average pooling is critical for  the relational inference performance.

\paragraph{Additional Inferred Relations in Basketball Trajectories}
We set the number of relations in Basketball dataset as two. In Sec 4.2, we visualized one of the inferred relations. Table \ref{fig:app_basketball}  visualizes the second relations. Notice that the first relation  captures focus on the rightmost red player while here the relation  captures focus on the leftmost red player.

\label{sec:app}

\end{document}


\maketitle

\section{Model Details}
In this section, we describe some details about the model implementation, especially the encoder part. Our encoder is analogical to the CNNs~\cite{he2016deep} used widely in the field of image recognition, where the task can be abstracted to a classification problem on 1D dimension. Meanwhile, inspired by~\cite{zhao2017pyramid}, we add additional 2 global feature extractors to efficiently combine the long-sequence (global) features  and the sub-sequence (local) features.

for ~\ours{}-C, Each residual block shown in Fig.3 consists of 4 skip-connection structures and for ~\ours{}-S, each residual block contains 2 skip-connection structures.

\section{Dataset Details}
Both of the two datasets consist of 50000 training examples, 10000 validation examples and 10000 testing examples. The best model is obtained by maximizing the ELBO on validation dataset and we report the evaluation results on the test dataset.
\paragraph{Particle dataset details}
In general, we use the same pre-processing in \nri{}. Each raw simulated trajectory has length of 4000 and we sample with frequency of 100 so that each sample has length of 40 in our dataset. Correspondingly, the unit of dynamic period/inference period matches the length of sample in our dataset. For instance, dynamic period = $10$ means that in the raw trajectory, the state of a node flips every 1000 time steps given sample frequency is 100. At each time step, the raw features of a particle consist of coordinates and velocities on $x$ and $y$ axes respectively. In addition, The value of the features are all normalized to range of $[0,1]$ and the evaluation is done on the same range as well. 
\paragraph{Basketball dataset details} The basketball dataset consists of trajectory from 30 teams in NBA. Each raw trajectory is captured with frequency of 25 ms. For our experiment, we sample the trajectories with frequency of 50 ms for more evident changes of the coordinates. Meanwhile, the values of the trajectories are normalized to range [0,1] as well and inference period matches the length of sample as well. For instance, inference period = $10$ means that our model produce prediction every 500 ms. At each time step, the raw features of a player consist of the coordinates on $x$ and $y$ axes respectively.

\section{Correction of Table 3 in the main paper.}
In Table 3, we made a mistake on the position of inference period and prediction period, thus the Table 3 in the paper wrongly presents our result. The correct version is shown in Table.~\ref{tab:correction} below.

\begin{table}
\centering
\caption{Correct version of Table 3}
\label{tab:correction}
\begin{tabular}{c|c|cccc}
\toprule
\multicolumn{2}{c|}{\textbf{Dynamic Period}} & 40 & 20 & 10 & 5 \\ \hline
\multirow{4}{*}{\begin{tabular}[c]{@{}c@{}}\textbf{Inference} \\ \textbf{Period} \end{tabular}}
 & 40(static) & \textbf{0.99} & 0.50 & 0.50 & 0.50 \\ 
 & 20         & 0.95 & \textbf{0.92} & 0.50 & 0.50 \\ 
 & 10         & 0.62 & 0.70 & \textbf{0.87} & 0.50  \\ 
 & 5          & 0.58 & 0.60 & 0.55 & \textbf{0.53} \\
\bottomrule
\end{tabular}
\end{table}

\section{Visualization of the other edge type in the Basketball Dataset}
In Sec 4.2, we visualized one of the two inferred edge types. And the other edge type is visualized in Fig.~\ref{fig:basketball_other}. Notice that the first edge type captures focus on the rightmost red player while the second edge type captures focus on the leftmost red player.

\begin{figure}[t!]
    \centering
    \begin{minipage}[c]{0.24\textwidth}
    \includegraphics[width=\textwidth]{figs/p50_pr.9_1_same.png_crop.png}
    \centering
    \label{fig:basketball}
    \end{minipage} 
    \hfill 	
    \begin{minipage}[c]{0.24\textwidth}
    \includegraphics[width=\textwidth]{figs/p50_pr.19_1_same.png_crop.png}
    \centering
    \label{fig:basketball}
    \end{minipage} 
    \hfill 	
    \begin{minipage}[c]{0.24\textwidth}
    \centering
    \includegraphics[width=\textwidth]{figs/p50_pr.29_1_same.png_crop.png}
    \label{fig:basketball}
    \end{minipage}
    \begin{minipage}[c]{0.24\textwidth}
    \includegraphics[width=\textwidth]{figs/p50_pr.39_1_same.png_crop.png}
    \centering
    \label{fig:basketball}
    \end{minipage}
    \hfill
    \begin{minipage}[c]{0.24\textwidth}
    \centering
    \includegraphics[width=\textwidth]{figs/p50_pr.9_1_cross.png_crop.png}
    \label{fig:basketball}
    \end{minipage} 
    \hfill 	
    \begin{minipage}[c]{0.24\textwidth}
    \centering
    \includegraphics[width=\textwidth]{figs/p50_pr.19_1_cross.png_crop.png}
    \label{fig:basketball}
    \end{minipage} 
    \hfill 	
    \begin{minipage}[c]{0.24\textwidth}
    \centering
    \includegraphics[width=\textwidth]{figs/p50_pr.29_1_cross.png_crop.png}
    \label{fig:basketball}
    \end{minipage}
    \hfill
    \begin{minipage}[c]{0.24\textwidth}
    \centering
    \includegraphics[width=\textwidth]{figs/p50_pr.39_1_cross.png_crop.png}
    \label{fig:basketball}
    \end{minipage}
    \caption{Visualization of the basketball players trajectories with inference period = 10. The top  row visualizes the inferred interactions from the same team (coordination) and the bottom row visualizes the inferred interactions from different teams (competition). Different columns represent different time steps. }
    \label{fig:basketball_other}
\end{figure}

\small
\bibliographystyle{ieee}
\bibliography{ref}